\documentclass[10pt,twocolumn,letterpaper]{article}

\usepackage{iccv}
\usepackage{times}
\usepackage{epsfig}
\usepackage{graphicx}
\usepackage{amsmath}
\usepackage{amssymb}
\usepackage{soul}
\usepackage{multirow}
\usepackage{multicol}

\usepackage{xcolor}

\newcommand{\myFigRef}[1]{Fig.~\ref{#1}}
\newcommand{\mySecRef}[1]{Sec.~\ref{#1}}

\definecolor{myPurple}{rgb}{0.4, .0, .8}
\definecolor{myGreen}{rgb}{0, .8, .3}
\definecolor{myRed}{rgb}{0.8, .2, .2}
\definecolor{myOrange}{rgb}{0.7, 0.45, 0.2}
\definecolor{myBlue}{rgb}{.0, .0, 1.0}
\definecolor{myBlue2}{rgb}{.0, .0, 0.5}
\definecolor{myBlack}{rgb}{.0, .0, 0.0}

\usepackage[breaklinks=true,bookmarks=false]{hyperref}

\iccvfinalcopy 

\def\iccvPaperID{2490} 

\ificcvfinal\fi

\begin{document}
\title{ PanoContext-Former: Panoramic Total Scene Understanding with \\ a Transformer}

\author{Yuan Dong$^{1}$, Chuan Fang$^{2}$, Liefeng Bo$^{1}$, Zilong Dong$^{1}$, Ping Tan$^{2}$ 
\thanks{$^{1}$ Alibaba Group, Hangzhou, China. Email:{\tt\small fangchuan.fc@alibaba-inc.com} $^{2}$ Hong Kong University of Science and Technology, Hong Kong. Email:{\tt\small cfangac@connect.ust.hk}}
}

\maketitle

\graphicspath{./figures/}

\begin{abstract}

Panoramic image enables deeper understanding and more holistic perception of $360^\circ$ surrounding environment, which can naturally encode enriched scene context information compared to standard perspective image. Previous work has made lots of effort to solve the scene understanding task in a bottom-up form, thus each sub-task is processed separately and few correlations are explored in this procedure. In this paper, we propose a novel method using depth prior for holistic indoor scene understanding which recovers the objects' shapes, oriented bounding boxes and the 3D room layout simultaneously from a single panorama. In order to fully utilize the rich context information, we design a transformer-based context module to predict the representation and relationship among each component of the scene.  In addition, we introduce a real-world dataset for scene understanding, including photo-realistic panoramas, high-fidelity depth images, accurately annotated room layouts, and oriented object bounding boxes and shapes.  Experiments on the synthetic and real-world datasets demonstrate that our method outperforms previous panoramic scene understanding methods in terms of both layout estimation and 3D object detection. 

\end{abstract}
\section{Introduction}
    Single-view indoor scene understanding from a single RGB image is  an essential yet challenging problem and has important applications such as augmented reality and service robotics. Most of the existing works solve room layout estimation, object detection, and reconstruction separately. Some recent works, including CooP~\cite{huang2018cooperative}, Total3D~\cite{nie2020total3dunderstanding}, and IM3D\cite{zhang2021holistic}, show that learning these tasks jointly helps to improve the performance on each subtask by exploiting context information. 
    In addition, panoramic image with a $360^\circ$ field-of-view (FOV) contains much richer information than a regular perspective image, whose FOV is normally around $60^\circ$. PanoContext~\cite{zhang2014panocontext} and DeepPanoContext~\cite{zhang2021deeppanocontext} prove that the context becomes significantly more robust and powerful with a larger FOV, which further improves the performance and enables accurate holistic scene understanding. Despite recent progress, the indoor scene understanding problem remains challenging since predicting object pose and shape from a single RGB image can be ambiguous without any 3D prior information in a real indoor environment with occlusion and clutter.
      
    \begin{figure}
    \includegraphics[width=\linewidth]{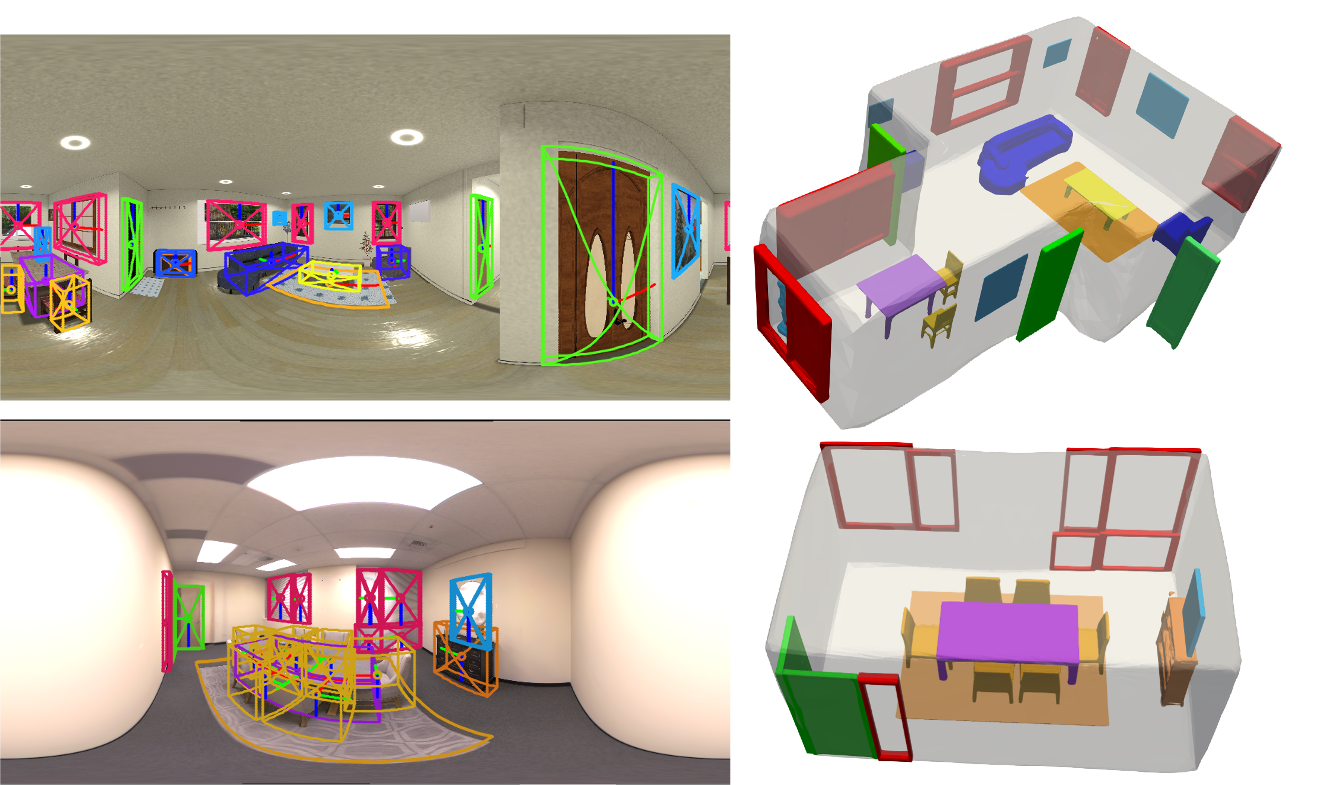}
    \caption{Given a single RGB panorama, we simultaneously estimate the room layout, oriented object bounding boxes (left), and full scene meshes (right). The first and second rows are examples from the iGibson-Synthetic~\cite{zhang2021deeppanocontext} and ReplicaPano datasets.}
    \label{fig:demostration}
    \end{figure}

    This paper proposes a new method for end-to-end total 3D scene understanding from a single panorama (~\myFigRef{fig:demostration}). Our approach has two important features. Firstly, we incorporate a monocular depth estimation sub-model to exploit 3D information to facilitate indoor scene understanding tasks. In this way, a point cloud based 3D object detector can be naturally applied to predict not only the 3D object boxes with semantic category labels but also the object shape codes. Our experiments show that integrating the estimated depth as a prior in a scene understanding framework can boost performance remarkably. We learn shape codes using an encoder that maps an object shape into an embedding representation, and then a decoder is used to recover the 3D shape of an object given its embedding vector. The observation is that the object features that are used to estimate boxes should contain information on object geometries; therefore, it is unnecessary to add an additional sub-model to predict object mesh. 
 
    
    \par
    Secondly, in order to better capture the global context in the scene, we unify different tasks together and propose a novel transformer-based context model for simultaneously predicting object shapes, oriented bounding boxes and 3D room layout. The key idea of this context model is to take all tokens as input to compute features for each task, in which the contribution of each token can be learned automatically by the attention mechanism. In addition, we also employ physical violation loss and random token masking strategy to strengthen the interactions across objects and room layout. Based on this idea, this model learns to discover context information among object-object and object-layout.
    
    \par
    When it comes to the panoramic datasets for holistic scene understanding, more efforts should be put into this area. Existing panoramic datasets are either for single application~\cite{xiao2012recognizing,zioulis2018omnidepth,yang2019dula,wang2018self} or missing critic 3D ground truth such as object boxes~\cite{chang2017matterport3d,armeni2017joint} and object shapes~\cite{zheng2020structured3d, zhang2014panocontext}. Compared with annotating the oriented object boxes and 3D shapes which is extremely labor-costing, it could be easier to generate ground truth from a simulator. Zhang~et~al.~\cite{zhang2021deeppanocontext} release the first holistic panoramic scene dataset with complete ground truth, rendering from synthetic scenes, while the panoramas lack realism and may set the barrier to deploy the algorithm into real-world. To minimize the domain gap between synthetic and real data, we render gravity aligned panoramas and depth images based on high-fidelity scene scan~\cite{straub2019replica}, then label layout, 3D object boxes and shapes accurately. 
    
    \par
    In general, the main contributions of our work can be summarized as follows:
    \begin{itemize}
        \item We propose a new method using depth prior for simultaneously estimating object bounding boxes, shapes, and 3D room layout from a single RGB panorama, followed by a novel transformer-based context model. To our best knowledge, it is the first work using a transformer to enable the network to capture context information efficiently for holistic 3D scene understanding.
        \item We introduce ReplicaPano, a real-world panoramic dataset comprising oriented bounding boxes, room layouts, and object meshes for panoramic 3D scene understanding.
        \item  The proposed method achieves state-of-the-art performance on both synthetic and real-world datasets. 
    \end{itemize}

   
\section{Related work}
\label{sec:related}
 \begin{figure*}[ht]
\centering
\includegraphics[width=\textwidth]{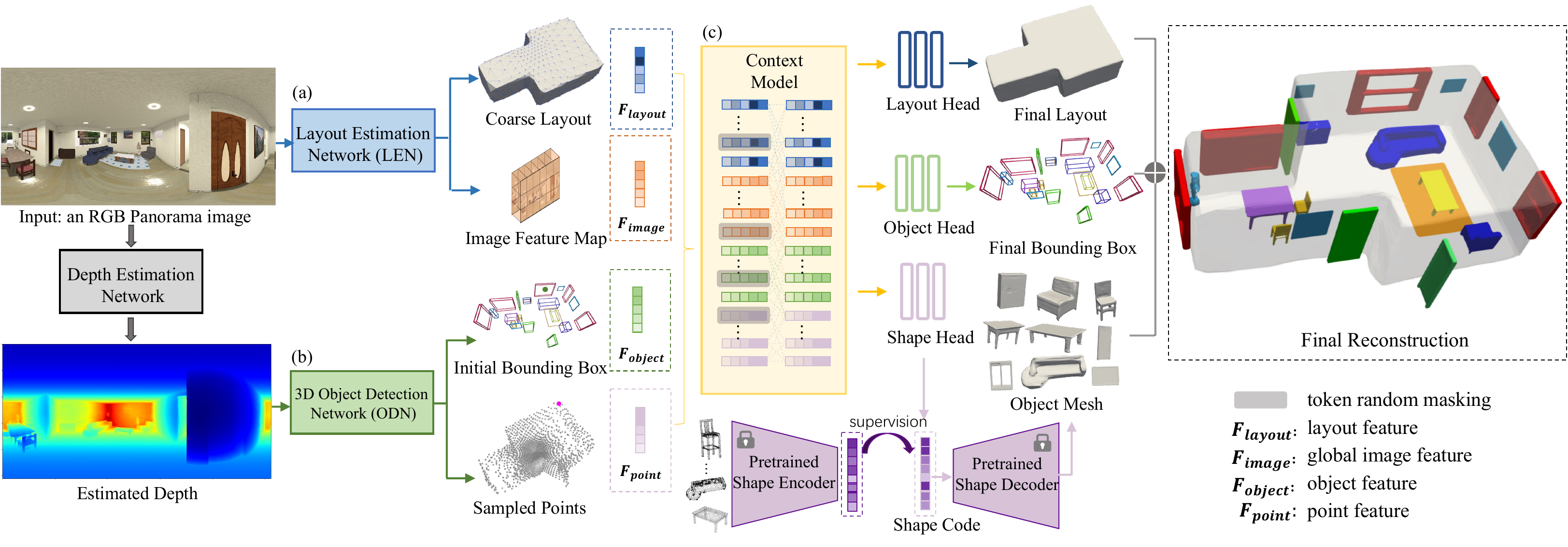}
\caption{The framework of the proposed holistic scene understanding pipeline. (a) The LEN module maps an panorama to a watertight 3D mesh of the room layout. (b) The ODN module jointly solves the oriented object bounding box and shape based on the estimated depth map of the indoor scene. (c) The Context module integrates various embeddings from LEN and ODN modules to fully explore the relationship among each component of the scene. Finally, refined features go through different heads, and the layout, oriented object bounding boxes, and shapes are recovered to reconstruct the full scene. }
\label{fig:system_framework}
\end{figure*}

\noindent\textbf{Single-View Scene Understanding} Scene understanding from a single image is highly ill-posed and ambiguous because of the unknown scale and severe occlusion in the scene. 
Many works have been proposed to study room layout estimation, 3D object detection and pose estimation, and 3D object reconstruction. Early room layout estimation works often make cuboid assumption~\cite{dasgupta2016delay, hedau2009recovering, lee2009geometric, mallya2015learning} or Manhattan assumption~\cite{zou2018layoutnet, yang2019dula, sun2019horizonnet, sun2021hohonet,wang2021led2}, while Pintore~et~al.~\cite{pintore2021deep3dlayout} model room structure as a 3D mesh to exploit the  possibility of estimating arbitrary room layout. 
Object detection works~\cite{huang2018cooperative, du2018learning,chen2019holistic++, tremblay2018deep} aim to infer 3D bounding boxes and object poses from 2D representation, with a 2D object detection~\cite{dai2017deformable, he2017mask} stage. 
In terms of object reconstruction, CAD models are selected from a large dataset to match the 2D object proposals in~\cite{huang2018holistic,izadinia2017im2cad,hueting2017seethrough}, while~\cite{chen2019learning,groueix2018papier,park2019deepsdf,mescheder2019occupancy,peng2020convolutional} demonstrated that implicit neural representations outperform grid,  point, and mesh-based representations in parameterizing geometry and seamlessly allow for learning priors over shapes.

Some recent works start to solve multiple tasks together to exploit context information. CooP~\cite{huang2018cooperative} introduces the target parameterizing and cooperative training scheme to solve for object poses and the layout of the indoor scene, but object shapes are absent. 
Total3D~\cite{nie2020total3dunderstanding} is the first work to solve layout, 3D object detection and pose estimation, and object reconstruction jointly. Zhang~et~al.~\cite{zhang2021holistic} proposes to improve the performance of all three tasks via implicit neural functions and graph convolutional networks. Liu~et~al.~\cite{liu2022towards} further improves the visual quality of indoor scene reconstruction using implicit representation. All these aforementioned methods only work on the perspective images, which lack enough information to better parse the entire scene. Zhang~et~al.~\cite{zhang2014panocontext} first introduced to parse indoor scenes using $360^\circ$ full-view panorama. Then, the follow-up work~\cite{zhang2021deeppanocontext} utilizes a deep learning-based framework that leverages image information and scene context to estimate objects' shapes, 3D poses and the room layout from a single panorama. Instead, we propose to incorporate depth prior and design a transformer-based context module for the panoramic scene understanding task, which can fully explore spatial context information among different components in an indoor scene.

\noindent\textbf{Transformer} Transformer~\cite{vaswani2017attention,devlin2018bert,liu2019roberta} has been the dominant network in the field of NLP for a few years. Inspired by ViT~\cite{dosovitskiy2020image}, researchers have designed many efficient networks~\cite{touvron2021training,yuan2021incorporating,wu2021cvt,liu2021swin,carion2020end} to combine the advantages of both CNNs and transformers. 


The review~\cite{xu2022multimodal} shows that the transformer structure can better learn context information among multi-modal input data.
CLIP~\cite{radford2021learning} jointly trains the image encoder and text encoder at the pretraining stage and converts an image classification task as a text retrieval task at test time.
Hu and Singh~\cite{hu2021unit} combined image and text to conduct multi-modal multi-task training and achieved good results in 7 visual and text tasks. Liu~et~al.~\cite{liu2021group} utilize the attention mechanism in transformers to fuse the object features and point features iteratively to generate more accurate object detection results from a point cloud. Similarly, conditional object query was used in ~\cite{wang2022bridged} to fuse point cloud and image features to obtain better results on the 3D detection task. There is a notable advantage of transformers for multi-modal tasks, in this paper, we introduce a transformer-based context module to facilitate holistic indoor scene understanding.


\noindent\textbf{Panoramic Dataset} SUN360~\cite{xiao2012recognizing} is the first real-world panoramic dataset used for place recognization, then it is annotated by Zhang~et~al.~\cite{zhang2014panocontext} for indoor scene understanding, but only room layout and objects' axis aligned bounding box are provided. 2D-3D-S~\cite{armeni2017joint} and Matterport3D~\cite{chang2017matterport3d} are published concurrently with real-world panoramas, but oriented object boxes and meshes are absent. 
In addition, there are some datasets~\cite{wang2018self,zioulis2018omnidepth,yang2019dula} published recently for the purpose of depth estimation or layout estimation on panorama. 
Zheng~et~al.~\cite{zheng2020structured3d} propose a large photo-realistic panoramic dataset for structured 3D modeling, namely Structured3D, but both mesh ground truths of scenes and objects are not published. To tackle that there is no panorama dataset with complete ground truths, author in~\cite{zhang2021deeppanocontext} uses iGibson~\cite{shen2021igibson} to synthesize 1500 panoramas with detailed 3D shapes, poses, semantics as well as room layout. However, the real-world panoramic indoor scene dataset containing all ground truth is still missing. To minimize the gap between synthetic and real-world data, we introduce a panoramic dataset rendered from real-scan~\cite{straub2019replica}, containing 2,700 photo-realistic panorama and high-fidelity depth images, accurately annotated room layout, and object bounding boxes and shapes. 
To our best knowledge, it's the first real-world image dataset with full ground truth for holistic scene understanding.

\section{Our Method}
\label{sec:method}
\par
The proposed pipeline simultaneously predicts the room layout, 3D object bounding boxes, and shapes with a depth estimation sub-model. As shown in ~\myFigRef{fig:system_framework}, we first estimate the whole-room depth map from the input panorama to facilitate the following modules. And the depth map will be converted into a point cloud, which can be used in the Object Detection Network (ODN) to jointly predict 3D object boxes and shape codes.
In the meantime, the layout is recovered as a triangle mesh from a single panorama through the Layout Estimation Network (LEN). 
In this paper, we exploit the transformer's intrinsic advantages and scalability in modeling different modalities and tasks, making it easier to learn appropriate spatial relationships among objects and layout. Features from layout, image, and 3D objects are fed into the context model to better estimate representations and relations among objects and layout. Finally, the room layout and object shapes are recovered as mesh, then scaled and placed into appropriate locations to reconstruct the full scene. We elaborate on the details of each module in this section.

\subsection{Layout Estimation}
\label{sec::method::LENModule}
As we want to relax the geometrical constraints applied to the output layout model (e.g., forcing vertical walls and/or planar walls and ceilings), we follow Pintore~et.~al.~\cite{pintore2021deep3dlayout} to map panoramic image to a triangle mesh representation $(V, E, F)$, where $V(n,3)$ is the set of $n=642$ vertices, $E(m,2)$ is the set of $m$ edges, each connecting two vertices, and $F(n,d)$ are the image feature vectors of dimension $d=288$ associated to vertices, denoted as $\textbf{F}_{\textbf{layout}}$ in the following.  Two Graph Convolution Network(GCN) blocks deform an initial tessellated sphere by offsetting its vertices, driven by associating image features to mesh vertices in a coarse-to-fine form.
Unlike~\cite{pintore2021deep3dlayout} only extracts features from equirectangular view, we additionally extract features from perspective views (e.g., ceiling and flooring views) through Equirectangular-to-Perspective (E2P) conversion. Then, E2P-based feature fusion~\cite{yang2019dula} is employed to fuse two types of features and get gravity aligned features. Specifically, we use ResNet-18 as the architecture for both equirectangular view and perspective views, the input dimension of image $\textbf{I}$ is $3 \times 512 \times 1024$, the output dimension of fused global image feature $\textbf{F}_{\textbf{image}}$ is $ 512 \times 16 \times 32$. The ablation experiment in ~\mySecRef{sec::Exp::Ablation} shows that the accuracy of room layout benefits from perspective features.

Drawing on the previous multi-modal
transformer models~\cite{lin2021end,lin2021mesh}, in order to fully associate the layout feature with the image feature, we inject the global image feature $\textbf{F}_{\textbf{image}}$ and layout features $\textbf{F}_{\textbf{layout}}$ from the first GCN block into the Context module, which will be elaborated in ~\mySecRef{sec::method::ContextModule}. Then the refined layout representation is sent into the layout head (the second GCN block). As a result, the second block returns the final deformed vertices $V^{*}(4n-6,3)$. 

\subsection{3D Object Detection and Mesh Generation}
\label{sec::method::ODNModule}
Our ODN adopts a similar structure of Group-Free~\cite{liu2021group} to accurately detect 3D objects in a point cloud. We first employ Unifuse~\cite{jiang2021unifuse} as our panoramic depth estimation network to generate a spherical depth map of the scene, then convert it into the form of a dense point cloud and rapidly sampled through Fabinacci Sampling. The following steps are the same as~\cite{liu2021group}, feeding the downsampled point cloud $S \in \mathbb R^{N \times 3}$ into the backbone network and the Initial Object Sampling module to get point cloud features and $K$ initial object candidates denoted as $\textbf{F}_{\textbf{point}} \in \mathbb R^{d_o \times M}$ and $\textbf{F}_{\textbf{object}} \in \mathbb{R}^{d_o \times K}$ respectively, where $K=256, M=1024$ and the feature dimension $d_o=288$. To automatically learn the contribution of all points to each object, these intermediate results will serve as points tokens and object tokens in the next subsection.  

Inspired by~\cite{irshad2022centersnap, irshad2022shapo}, we observe that shape information is embedded in the object feature in the process of 3D object detection. Thus, in addition to the existing object prediction head,  we add a shape prediction head to jointly predict the shape latent code and bounding box of the candidate object. The shape latent code is supervised by a pretrained autoencoder of object mesh, here we choose ONet~\cite{mescheder2019occupancy} to serve as the autoencoder, because of its computation-friendly size of object shape latent code (1D vector of size 512 ), which can be easily used to construct the shape loss during the training. The ONet is pre-trained on ShapeNet~\cite{chang2015shapenet} and refined on iGibson-Synthetic~\cite{zhang2021deeppanocontext} with data augmentation.

\subsection{Transformer-based Context Module}
\label{sec::method::ContextModule}
\begin{figure}
    \centering
    \includegraphics[width=\linewidth, height=0.5\textwidth]{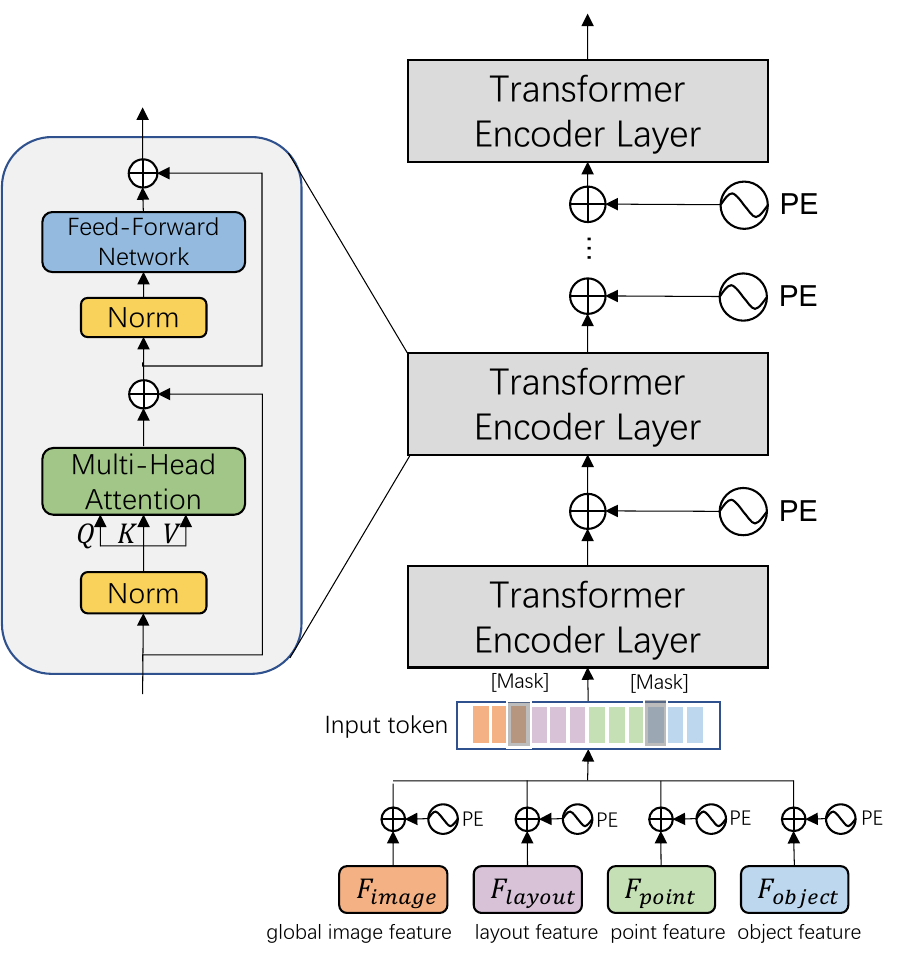}
    \caption{Architecture of the Context module. }
    \label{fig:context_module}
\end{figure}

Given a single panorama, our goal is to further explore the intrinsic relationships among different components of the indoor scene. We designed the transformer-based context module with a multi-layer encoder structure to extract better representations of objects and room layouts from different features. As shown in~\myFigRef{fig:context_module}, the position embeddings of point, object, layout, and global image are computed by applying independent linear layers on the parameterization vector of point $(x, y, z)$, 3D box $(x, y, z, l, h, w)$, layout vertice $(x,y,z)$, and unit spherical coordinate $(cos\phi sin\theta,sin\phi,cos\phi cos\theta)$, respectively. The global image feature $\textbf{F}_{\textbf{image}}$ along with point feature $\textbf{F}_{\textbf{point}}$, object feature $\textbf{F}_{\textbf{object}}$, and layout feature $\textbf{F}_{\textbf{layout}}$ are point-wise summed with their position embeddings and then are concatenated together and act as the input for the context module:
\begin{equation}
     \mathbf Z=[\textbf{F}_{\textbf{image}}, \textbf{F}_{\textbf{layout}}, \textbf{F}_{\textbf{point}}, \textbf{F}_{\textbf{object}}]. 
\label{eq::context_input}
\end{equation}

The context module is composed of 6 stacked transformer encoder layers, each layer includes a multi-head self-attention (MHSA) layer and a feed-forward network (FFN). MHSA is the foundation of a transformer, allowing the model to jointly attend to information from different representation subspaces. In a self-attention module, embedding $\mathbf Z$ will go through three projection matrices ($\mathbf W^{Q}$, $\mathbf W^{K}$, $\mathbf W^{V}$) to generate three embedding $\mathbf Q$(query), $\mathbf K$(key) and $\mathbf V$(value):
\begin{equation}
        \mathbf{Q=ZW^Q}, \mathbf{K=ZW^K}, \mathbf{V=ZW^V}.
\label{eq:self_attention_projecting}
\end{equation}
The output of self-attention is the aggregation of the values that are weighted by the attention weights.
In our case, we propose a token random masking scheme to help the encoder to be robust and effective in handling situations with heavy occlusions, formulated as:
\begin{equation}
   {\rm MSA}(\mathbf Q, \mathbf K, \mathbf V) =
    softmax \left( \frac{\mathbf{Q} \mathbf{K}^T}{\sqrt{d}} \odot \mathbf{M} \right) \mathbf{V}, \\
\label{eq:mask_self_attention}
\end{equation}
where $d$ is the dimension of query embedding and $\mathbf{M}$ is the specific masking matrix. Multiple self-attention layers are stacked and their concatenated outputs are fused by weighting matrix $\mathbf{W^h}$, to form MHSA:
\begin{equation}
   {\rm MHSA}(\mathbf Q, \mathbf K, \mathbf V) = \sum_{h=1}^{H} {\rm MSA}(\mathbf Q, \mathbf K, \mathbf V)\mathbf{W^h}.
\label{eq:mhsa}
\end{equation}
After iterative refinement of MHSA, the resulting embedding of different stages are fed into different prediction heads to generate the results of each task, which will be ensemble to produce superior results.

\subsection{Loss Function}
\label{sec::method::LossFunction}
In this section, we conclude the learning targets with the corresponding loss functions, and describe our joint loss for end-to-end training.

\noindent\textbf{Layout Loss} At first, we adopt the loss function from Pintore~et~al.~\cite{pintore2021deep3dlayout} to define layout loss which measures the prediction with respect to the ground truth layout:
\begin{equation}
    \mathcal{L}_{layout} = \lambda_p * \mathcal{L}_{pos} + \lambda_n * \mathcal{L}_{norm} +\lambda_e * \mathcal{L}_{sharp}.
\label{eq::loss::layout}
\end{equation}
where $\mathcal{L}_*$ and $\lambda_*$ are the losses and coefficients for vertex position, surface normal, and edge sharpness, respectively.

    \noindent\textbf{Object Loss} The loss for ODN is similar to~\cite{liu2021group}, including sampling loss $\mathcal{L}_{\rm samp}$, objectness loss $\mathcal{L}_{\rm objness}$, classification loss $\mathcal{L}_{\rm cls}$, center offset loss $\mathcal{L}_{\rm cen}$, size classification loss $\mathcal{L}_{\rm {size\_cls}}$, and size offset loss $\mathcal{L}_{\rm {size\_off}}$. Additionally, $\left . 1 \right )$ since we aim to estimate the orientated bounding box of the object, the box's heading prediction with a cross-entropy loss $\mathcal{L}_{\rm {head\_cls}}$ and a smooth-L1 loss $\mathcal{L}_{\rm {head\_off}}$ is included. $\left . 2 \right )$ the shape code prediction loss $\mathcal{L}_{\rm shape}$. Let $\theta$ denote the estimated shape codes, we use a smooth-L1 loss to minimize the
errors between predictions and ground truth:
\begin{equation}
    \mathcal{L}_{\rm shape} = \frac{1}{K} \sum_{k=1}^{K} {\ell_1 {\left (\theta - \bar{\theta} \right )}}, 
\label{eq::loss::object_shape}
\end{equation}
where  ground truths $\bar{\theta}$ are given from pre-trained autoencoder.
For the sake of brevity, these losses will be referred to as a set $\{\mathcal{L}_{object\_loss}\}$. 

We define the object estimation losses on all encoder layers in the context module, which are averaged to form the final loss:
\begin{equation}
\begin{split}
    \mathcal{L}_{object} =& \frac{1}{L} \sum_{l=1}^{L} \mathcal{L}_{obj}^{l}, \\
    \mathcal{L}_{obj}^{l} =& \sum_{x \in \{\mathcal{L}_{object\_loss}\} } {\beta_x * \mathcal{L}_{x}}.
\end{split}
\label{eq::loss::object}
\end{equation}
Each $\beta_x$ is the loss weight corresponding to the specific object loss.

\begin{figure}
    \centering
    \includegraphics[width=\linewidth]{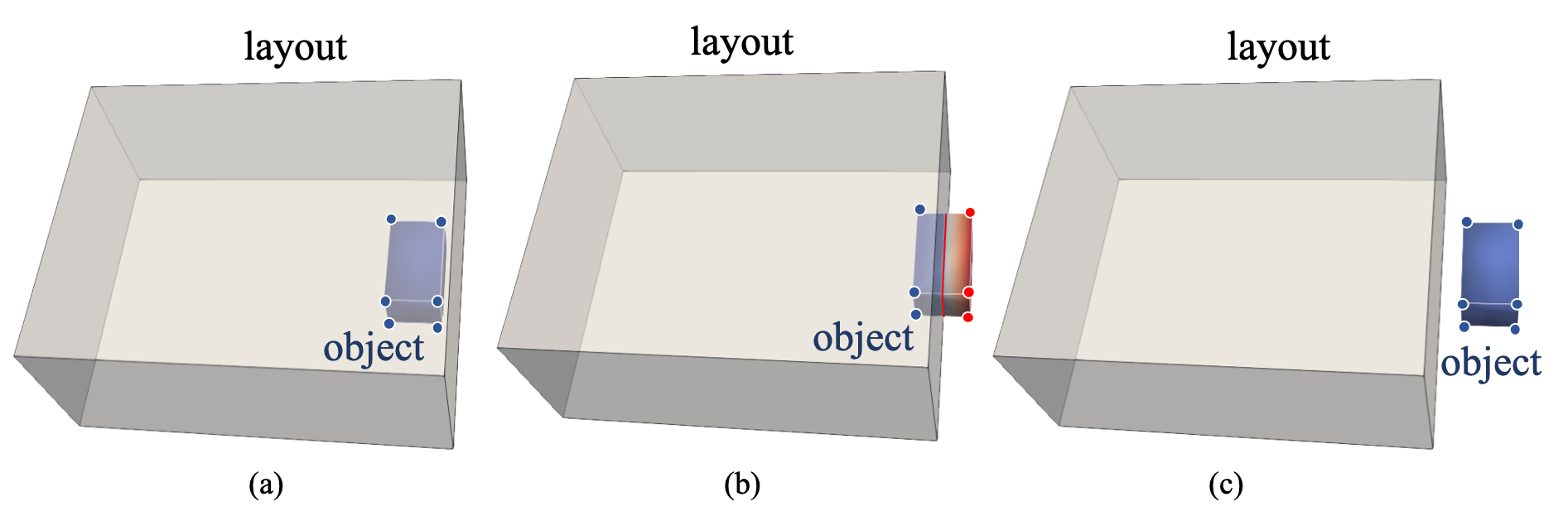}
    \caption{Object-Layout physical violation example. The physical violation loss is calculated only when the object intersection with layout (b). There is no physical constraint when the object is completely inside (a) or outside the layout (c).  }
    \label{fig:physical_violation_loss}
\end{figure}

\noindent\textbf{Physical Violation Loss} In order to produce a physically plausible scene and regularize the relationships between objects and layout, we add physical violation loss as a part of the joint loss. As shown in ~\myFigRef{fig:physical_violation_loss}, when the bounding box of an object intersects with the layout (i.e., walls, ceiling, or floor), the physical violation loss is calculated with the Manhattan distance to the layout. Some types of objects do intersect with the layout, such as windows and doors. So the physical constraints are only applied for categories that should never intersect with the layout. 
The physical violation loss is defined as:
\begin{equation}
\begin{split}
    \mathcal{L}_{physic} =& \frac{1}{K} \sum_{\substack{k=1}}^{K}  \mathds{1}_{ins}\  \mathcal{L}_{\rm {3d\_violation}}, \\
    \mathcal{L}_{\rm {3d\_violation}} = & \sum_{i=1}^{8}(relu(x_{i}^{k}  - max(X^L)) \\ 
    &+  relu(min(X^L)-x_{i}^{k})) ,
\end{split}
\label{eq::loss:physical_violation_loss}
\end{equation}
where $x_{i}^{k}$ is corner of the $k$th object bounding box, $\rm X^L$ is a set of vertices of layout mesh. The $relu$ is applied to consider only the outside corners. $ \mathds{1}_{ins}$ has a value of 1 if the bounding box is not completely outside of the layout, and a value of 0 otherwise.

\noindent All the loss functions in joint training can be defined as :
\begin{equation}
    \mathcal{L} = \sigma_l * \mathcal{L}_{layout} + \sigma_o * \mathcal{L}_{object} + \sigma_p * \mathcal{L}_{physic}.
\end{equation}

\subsection{Panoramic Dataset}
\label{sec::method::Dataset}
For now, the realistic panoramic dataset with all ground truth is still missing. To benefit the community, we publish \textbf{ReplicaPano}, a real-world panoramic scene understanding dataset with full ground truth. With the help of the high-fidelity textured mesh provided by Replica dataset~\cite{straub2019replica}, we render photo-realistic panorama from 27 rooms diversely furnished by 3D objects. For each room, we randomly render 100 pairs of equirectangular RGB and depth images, all the images are gravity aligned and the height of the camera center is 1.6m. Given a panorama, we utilize PanoAnnotator~\cite{yang2018panoannotator} to accurately label the room layout. Based on the colored point cloud and semantic segmentation information provided by Replica, we semi-automatically annotate the bounding box for each object in each room. Following the  NYU-37 object labels~\cite{silberman2012indoor}, we select 25 categories of objects that are commonly seen in indoor scenes. Because the complete object mesh is not given in Replica, we look through large-scale 3D shape datasets ShapeNet~\cite{chang2015shapenet}, 3D-FUTURE~\cite{fu20213d}  and ReplicaCAD~\cite{szot2021habitat} to match the object observed in the image. Finally, we get 2,700 photo-realistic panoramas with depth images, room layouts, 3D object bounding boxes, and object meshes. More samples of ReplicaPano can be found in the supplementary files.

\section{Experiment}
\label{sec::Exp}

\begin{table*}[t]
\begin{center}
\setlength{\tabcolsep}{1.4mm}
\begin{tabular}{|l|c|c|c|c|c|c|c|c|c|c|c|c|}
\hline
Method & chair & soft & table & fridge & sink & door & \makecell[c]{floor\\lamp} & \makecell[c]{bottom\\cabinet} & \makecell[c]{top\\cabinet} & \makecell[c]{sofa\\chair} & dryer & mAP$\uparrow$\\ 
\hline
\hline
Total-Pano & 20.84 & 69.65 & 31.79 & 43.13 & 68.42 & 10.27 & 16.42 & 34.42 & 20.83 & 62.38 & 33.78 & 37.45 \\
Im3D-Pano & 33.08 & 72.15 & 37.43 & 70.45 & 75.20 & 11.58 & 6.06 & 43.28 & 18.99 & 78.46 & 41.02 & 44.34 \\
DeepPanoContext & 27.78 & 73.96 & 46.85 & 74.22 & 75.29 & 21.43 & 20.69 & 52.03 & 50.39 & 77.09 & 59.91 & 52.69 \\
DeepPanoContext-depth & \textbf{39.41} & 78.03 & 51.44 & 75.24 & 81.46 & 51.97 & \textbf{60.01} & 55.56 & 42.58 & 79.99 & \textbf{60.07} & 61.43 \\
Group-Free & 27.83 & 96.04 & 61.57 & \textbf{84.69} & 87.69 & 82.20 & 27.20 & 56.46 & 77.99 & 79.21 & 8.29 & 62.65 \\
Ours & 38.47 & \textbf{98.15} & \textbf{66.61} & 82.77 &\textbf{89.55} & \textbf{87.49} & 40.31 & \textbf{59.53} & \textbf{80.71} & \textbf{83.42} & 13.83 & \textbf{67.35} \\
\hline
\end{tabular}
\end{center}
\caption{Comparisons of object detection on iGibson-Synthetic with state-of-the-art. We use mean average precisions with 3D IoU threshold 0.15 and evaluate 11 common object categories following~\cite{nie2020total3dunderstanding,zhang2021holistic,zhang2021deeppanocontext}. DeepPanoContext-depth is the extended version with depth map.}
\label{table:object_detection_iGibisn}
\end{table*}

\begin{table*}[t]
\begin{center}
\setlength{\tabcolsep}{1.9mm}
\begin{tabular}{|l|c|c|c|c|c|c|c|c|c|c|c|}
\hline
Method & cabinet & door & chair & curtain & lamp & rug & sofa & table & trash & tv & mAP$\uparrow$ \\
\hline\hline
DeepPanoContext & 35.33 & 6.78 & 47.04 & 13.6 & 12.15 & 4.49 & 26.87 & 73.34 & 39.59 & 4.86 & 26.41 \\
DeepPanoContext-depth & 52.49 & 11.42 & \textbf{70.39} & 32.38 & 20.02 & 9.10 & 30.13 & \textbf{82.24} & \textbf{63.22} & 12.19 & 38.36 \\
Group-Free & 59.56 & 42.21 & 52.83 & \textbf{34.07} & 19.65 & 32.90 & 80.59 & 51.47 & 44.64 & \textbf{52.76} & 47.07 \\
Ours & \textbf{63.69} & \textbf{46.74} & 54.02 & 30.41 & \textbf{20.04} & \textbf{48.53} & \textbf{80.96} & 46.42 & 51.53 & 47.82 & \textbf{49.02} \\
\hline
\end{tabular}
\end{center}
\caption{Comparisons of object detection on ReplicPano.}
\label{table:object_detection_replicapano}
\end{table*}

In this section, we compare our model with both holistic scene understanding and single-task methods and perform ablation studies to analyze the effectiveness of the key components.

\subsection{Experiment Setup}
\noindent \textbf{Dataset.} 
We use two panoramic datasets in our experiments. \textbf{1) iGibson-Synthetic.} The panoramic images are synthesized using the iGibson simulator~\cite{shen2021igibson}. Same as the setting in DeepPanoContext~\cite{zhang2021deeppanocontext}, we use 10 scenes for training and 5 scenes for testing. \textbf{2) ReplicaPano.} To demonstrate our work's efficiency in real-world scenes, among 27 rooms, we use 16 for training, 4 for validation, and 7 for testing.

\noindent\textbf{Metrics.} The results of each sub-task are evaluated with the metrics used in previous works~\cite{nie2020total3dunderstanding,zhang2021holistic,zhang2021deeppanocontext}. Object detection is measured using mean average precision (mAP) with the threshold of 3D bounding box IoU set at 0.15. The room layout estimation error is tested by standard metrics for indoor layout reconstruction (i.e., 2D-IoU and 3D-IoU) followed by Pintore et.al~\cite{sun2019horizonnet,sun2021hohonet,pintore2021deep3dlayout}. Since the object mesh generation in our method is significantly different from other scene understanding work, we only compare the result with that of others qualitatively.

\noindent\textbf{Implementation.} The borrowed monocular depth estimation network (i.e., Unifuse~\cite{jiang2021unifuse}) and 3D auto-encoder network (i.e., ONet~\cite{mescheder2019occupancy}) are finetuned individually on each dataset from the weights pretrained on Matterport3D and ShapeNet, respectively. The input point cloud for the object detection network is sampled to 50K by Fibonacci sampling from the estimated depth. The auto-encoder network takes 300 points from the surface of each watertight model as input and embeds each sample as a vector of size 512. In the context model, ten percent of tokens are randomly masked. We trained object detection, layout estimation, and mesh generation jointly with randomly initialized parameters on a single NVIDIA V100 GPU. More training details are given in the supplementary files.

 \begin{figure*}
\centering
\includegraphics[width=1\textwidth]{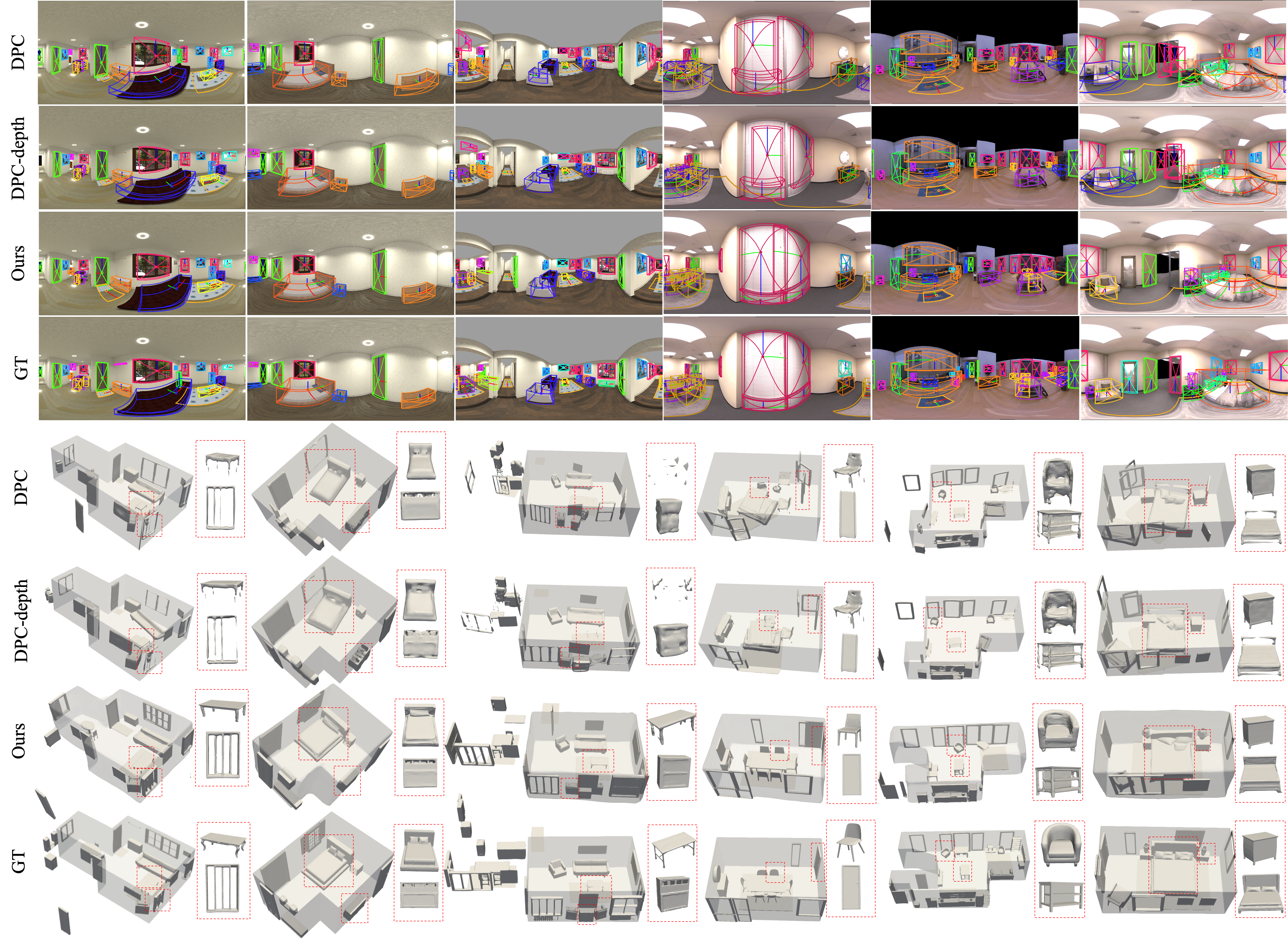}
\caption{Qualitative comparisons on 3D object detection and scene reconstruction. In the top four rows, we compare our object detection results with DeepPanoContext (DPC), DeepPanoContext with depth map (DPC-depth), and ground truth in the panoramic view. The color of the bounding boxes represents their categories. The bottom four rows show the results of scene reconstruction, with two magnified object reconstruction results presented on the right-hand side. Note that the first three columns are the results on iGibson-Synthetic, and the last three columns are the results on ReplicaPano.}
\label{fig:results}
\end{figure*}

\subsection{Comparisons with State-of-the-art Methods}
\noindent\textbf{Object Detection} We compare our 3D object detection results with previous state-of-the-art holistic scene understanding and single-task learning methods. DeepPanoContext~\cite{zhang2021deeppanocontext} is the only method to achieve total 3D scene understanding directly on panoramic image. Total3D~\cite{nie2020total3dunderstanding} and IM3D~\cite{zhang2021holistic} that work with perspective image are extended to panorama for comparison on iGibson-Synthetic dataset. In order to show the effectiveness of depth prior in the scene understanding task, We extend DeepPanoContext with an estimated depth map as follows: we use PointNet++ to extract the object geometry feature and concatenate this feature with other appearance features to estimate 3D bounding boxes. As for the single task comparison, the point-based object detection method Group-Free~\cite{liu2021group} is chosen as baseline. The results of each method on iGibson-Synthetic are shown in Tab.~\ref{table:object_detection_iGibisn}. 
Since DeepPanoContext shows higher performance than Total3D and IM3D, we only compare it and its extension on ReplicaPano, results can be found  in Tab.~~\ref{table:object_detection_replicapano}.

As shown in Tab.~\ref{table:object_detection_iGibisn} and Tab.~\ref{table:object_detection_replicapano}, our proposed method consistently outperforms both holistic understanding methods and the point-based detection baseline on most categories and the average mAP. We can see that DeepPanoContext has been significantly improved by integrating the estimated depth map, which indicates the depth prior is absolutely necessary. The table shows our method gains better results for categories that are closely related to room layout, such as door and rug, since the transformer-based context model encourages rational spatial relationships among objects and room layout. For a few categories such as floor lamp and chair, DeepPanoContext-depth performs better, the gap between these categories owns to two factors: 1) The depth estimation model failed to recover tiny structure, for example, the pole of a floor lamp, which deteriorates the performance of our method. 2) DeepPanoContext-depth uses a finetuned 2D detector to initialize the estimation and achieve good performance for heavily occluded objects (e.g., chairs are occluded behind a table). Improving depth quality and introducing a 2D detector into our method may help to improve the accuracy further.

\begin{table}[t]
\begin{center}
\setlength{\tabcolsep}{1.0mm}
\begin{tabular}{|l|cc|cc|}
\hline
\multirow{2}{*}{Method} & \multicolumn{2}{c|}{iGibson-Synthetic} & \multicolumn{2}{c|}{ReplicaPano} \\ \cline{2-5} 
 & \multicolumn{1}{c|}{2D-IoU$\uparrow$} & \multicolumn{1}{l|}{3D-IoU$\uparrow$} & \multicolumn{1}{c|}{2D-IoU$\uparrow$} & \multicolumn{1}{l|}{3D-IoU$\uparrow$} \\ 
\hline\hline
HorizionNet & \multicolumn{1}{c|}{89.22} & 89.18 & \multicolumn{1}{c|}{84.56} & 83.59 \\
HoHoNet & \multicolumn{1}{c|}{90.13} & 89.97 & \multicolumn{1}{c|}{84.76} & 84.05 \\
Led2Net & \multicolumn{1}{c|}{90.39} & 90.30 & \multicolumn{1}{c|}{84.62} & 83.91 \\
Deep3dLayout & \multicolumn{1}{c|}{90.65} & 90.40 & \multicolumn{1}{c|}{84.87} & 83.50 \\
Ours & \multicolumn{1}{c|}{\textbf{92.24}} & \textbf{92.04} & \multicolumn{1}{c|}{\textbf{85.98}} & \textbf{84.58} \\ \hline
\end{tabular}
\end{center}
\caption{Comparisons of layout estimation on iGibson-Synthetic and ReplicaPano. Evaluation metrics include 2D and 3D intersection-over-union (IoU) following~\cite{sun2019horizonnet,sun2021hohonet,pintore2021deep3dlayout}.}
\label{table:layout_estimation}
\end{table}

\noindent\textbf{Layout Estimation} Previous panoramic scene understanding work does not give quantitative analysis in terms of the layout estimation, thus we only compare our method with recent state-of-the-art layout estimation methods~\cite{sun2019horizonnet,sun2021hohonet,wang2021led2,pintore2021deep3dlayout}. As shown in Tab.~\ref{table:layout_estimation}, our method achieves the best performance among other baselines, indicating joint training with the context model helps to improve the layout estimation from a single panorama.

\noindent\textbf{Holistic Scene Reconstruction} Qualitative comparison with DeepPanoContext and DeepPanoContext-depth are demonstrated in Fig.~\ref{fig:results}, our method obtains the best indoor scene reconstruction, including the object pose, room layout, and object shape reconstruction.

\begin{table}[t]
\begin{center}
\setlength{\tabcolsep}{0.7mm}
\begin{tabular}{|l|cc|c|}
\hline
\multirow{2}{*}{Method} & \multicolumn{2}{c|}{depth metric} & detection metric \\ \cline{2-4} 
 & \multicolumn{1}{c|}{Abs.Rel.$\downarrow$} & RMSE$\downarrow$ & mAP$\uparrow$ \\ 
 \hline\hline
Panoformer-pretrain & \multicolumn{1}{c|}{0.0774} & 0.2105 & 54.77 \\
Unifuse-finetune & \multicolumn{1}{c|}{0.0328} & 0.1107 & 67.35 \\
Panoformer-finetune & \multicolumn{1}{c|}{0.0214} & 0.0997 & 69.09 \\
GT-depth & \multicolumn{1}{c|}{-} & - & 79.46 \\
\hline
\end{tabular}
\end{center}

\caption{The impact of depth accuracy. Evaluation metrics include absolute relative error (Abs. Rel.) and root mean square error (RMSE) for depth and mAP for object detection. Panoformer-pretrain is pre-trained on Matterport3D~\cite{chang2017matterport3d}, while *-finetune means the depth estimator gets finetuned on iGibson-Synthetic.}
\label{table:ablation_depth_error}
\end{table}

\begin{table*}[tp]
\begin{center}
\setlength{\tabcolsep}{1mm}
\begin{tabular}{|c|c|c|c|c|c|c|c|c|}
\hline
\makecell[c]{Perspective \\Feature} & \makecell[c]{Joint\\ Training} & \makecell[c]{Physical\\ Violation Loss} & \makecell[c]{Token\\ Masking} & \makecell[c]{Image\\ Token} & \makecell[c]{mAP$\uparrow$\\(11 categories)} & \makecell[c]{mAP$\uparrow$\\(57 categories)} & 2D-IoU$\uparrow$ & 3D-IoU$\uparrow$ \\
\hline\hline
{\xmark} & {\xmark} & {\xmark} & {\xmark} & {\xmark} & 62.59 & 40.44 & 90.65 & 90.40 \\
\ding{51} & {\xmark} & {\xmark} & {\xmark} & {\xmark} & - & - & 90.98 & 90.73 \\
\ding{51}&  \ding{51}&  \ding{51}& {\xmark} & {\xmark} & 65.68 & 41.50 & 91.41 & 91.20 \\
 \ding{51}&  \ding{51}&  \ding{51}&  \ding{51}& {\xmark} & 66.27 & 42.22 & 92.14 & 91.78 \\
 \ding{51}& \ding{51}&  \ding{51}& {\xmark} &  \ding{51}& 66.78 & 41.97 & 91.77 & 91.56 \\
 \ding{51}&  \ding{51}&  \ding{51}&  \ding{51}&  \ding{51}& \textbf{67.35} & \textbf{43.55} & \textbf{92.24} &\textbf{ 92.04} \\
\hline
\end{tabular}
\end{center}
\caption{The ablation studies on iGibson-Synthetic dataset, demonstrates how our proposed designs improve the accuracy on object detection and layout estimation.  We show in the last row the full architecture setup.}
\label{table:ablation_study}
\end{table*}

\subsection{Ablation Study}
\label{sec::Exp::Ablation}
In this section, we conduct some ablation studies on iGibson-Synthetic to clarify the importance of each component in our method.
\\\noindent\textbf{Impact of depth quality} We first investigate how the accuracy of the depth map impacts the final 3D object detection. Two depth estimation networks, Unifuse~\cite{jiang2021unifuse} and PanoFormer~\cite{shen2022panoformer} are involved in Tab.~\ref{table:ablation_depth_error}, which reveal that object detection results benefit from higher depth quality. In addition, we observe that even if the proposed method uses a depth estimator without finetuning, the performance still slightly outpasses that of DeepPanoContext (Tab.~\ref{table:object_detection_iGibisn}), which employed a 2D detector for initialization.
\\\noindent\textbf{Effect of architecture and loss} To figure out the effect of each module, we provide detailed ablation experiments in terms of object detection and layout estimation. The results are summarized in Tab.~\ref{table:ablation_study}. The first 2 rows show the room layout benefit from perspective features. The third row indicates that introducing joint training and physical violation loss consistently improves the results of object detection and layout estimation. As for the fourth and fifth rows, we can conclude that our method can generate better representation and relationships among objects and the room layout, with the help of global image tokens and the token masking strategy, thus obtaining better results on each task.

\section{Conclusion}
In this paper, we propose a new method for end-to-end 3D indoor scene understanding from a single RGB panoramic image with depth prior.
To better learn the context information in the panorama, we use a Transformer-based context model to learn the relationship between objects and room layout. In addition, we introduce a new real-world dataset for panoramic holistic scene understanding. Experiments demonstrate that our method achieves state-of-the-art performance on both synthetic and real-world datasets.

{\small
\bibliographystyle{ieee_fullname}
\bibliography{egbib.bib}
}

\end{document}